\documentclass{article}

\usepackage{times}
\usepackage{graphicx} 

\usepackage{natbib}

\usepackage{algorithm}
\usepackage{algorithmic}

\usepackage[nohyperref,accepted]{icml2017}

\icmltitlerunning{Semi-Supervised Generation with Cluster-aware Generative Models}

\usepackage{subcaption}
\usepackage{amsmath,amsfonts,amsthm,amssymb} 
\usepackage{wrapfig}
\usepackage[normalem]{ulem}
\usepackage{amsmath}
\usepackage[disable]{todonotes}
\usepackage{tikz}
\usepackage{xspace}
\usetikzlibrary{bayesnet}
\usetikzlibrary{arrows}
\usetikzlibrary{shapes.multipart}
\tikzset{
state/.style={
       rectangle split,
       rectangle split parts=2,
       rectangle split part fill={red!30,blue!20},
       rounded corners,
       draw=black, very thick,
       minimum height=2em,
       text width=3cm,
       inner sep=2pt,
       text centered,
       }
}

\def\namep{Cluster-aware Generative Models }
\def\nm{CaGeM\xspace}
\def\drm{\mathrm{d}}

\begin{document} 

\twocolumn[
\icmltitle{Semi-Supervised Generation with Cluster-aware Generative Models}



\icmlsetsymbol{equal}{*}

\begin{icmlauthorlist}
\icmlauthor{Lars Maal{\o}e}{dtu}
\icmlauthor{Marco Fraccaro}{dtu}
\icmlauthor{Ole Winther}{dtu}
\end{icmlauthorlist}

\icmlaffiliation{dtu}{Technical University of Denmark}
\icmlcorrespondingauthor{Lars Maal{\o}e}{larsma@dtu.dk}
\icmlcorrespondingauthor{Marco Fraccaro}{marfra@dtu.dk}
\icmlcorrespondingauthor{Ole Winther}{olwi@dtu.dk}

\icmlkeywords{deep learning, variational inference, variational auto-encoders, generative modeling}

\vskip 0.3in
]



\printAffiliationsAndNotice{}  

\begin{abstract} 
Deep generative models trained with large amounts of unlabelled data have proven to be powerful within the domain of unsupervised learning.
Many real life data sets contain a small amount of labelled data points, that are typically disregarded when training generative models.
We propose the \textit{Cluster-aware Generative Model}, that uses unlabelled information to infer a latent representation that models the natural clustering of the data, and additional labelled data points to refine this clustering. 
The generative performances of the model significantly improve when labelled information is exploited, obtaining a log-likelihood of $-79.38$ nats on permutation invariant MNIST, while also achieving competitive semi-supervised classification accuracies. The model can also be trained fully unsupervised, and still improve the log-likelihood performance with respect to related methods. 
\end{abstract} 

\section{Introduction}

Variational Auto-Encoders (VAE) \citep{Kingma13,Rezende14} and Generative Adversarial Networks (GAN) \citep{Goodfellow2014} have shown promising generative performances on data from complex high-dimensional distributions.
Both approaches have spawn numerous related deep generative models, not only to model data points like those in a large unlabelled training data set, but also for semi-supervised classification \citep{Kingma14,Maaloe2016,Springenberg2015,Salimans2016}. In semi-supervised classification a few points in the training data are endowed with class labels, and the plethora of unlabelled data aids to improve a supervised classification model. 

Could a few labelled training data points in turn improve a deep generative model? This reverse perspective, doing semi-supervised generation, is investigated in this work. Many of the real life data sets contain a small amount of labelled data, but incorporating this partial knowledge in the generative models is not straightforward, because of the risk of overfitting towards the labelled data. This overfitting can be avoided by finding a good scheme for updating the parameters, like the one introduced in the models for semi-supervised classification \citep{Kingma14,Maaloe2016}. However, there  is a difference in optimizing the model towards  optimal classification accuracy and generative performance. We introduce the \textit{Cluster-aware Generative Model (CaGeM)}, an extension of a VAE, that improves the generative performances, by being able to model the natural clustering in the higher feature representations through a discrete variable \citep{Bengio2013a}. The model can be trained fully unsupervised, but its performances can be further improved using labelled class information that helps in constructing well defined clusters. A generative model with added labelled data information may be seen as parallel to how humans rely on abstract domain knowledge in order to efficiently infer a causal model from property induction with very few labelled observations \citep{Tenenbaum2006}.

Supervised deep learning models with no stochastic units are able to learn multiple levels of feature abstraction. In VAEs, however, the addition of more stochastic layers is often accompanied with a built-in pruning effect so that the higher layers become disconnected and therefore not exploited by the model \citep{Burda15,Sonderby2016}. As we will see, in CaGeM the possibility of learning a representation in the higher stochastic layers that can model clusters in the data drastically reduces this issue. This results in a model that is able to disentangle some of the factors of variation in the data and that extracts a hierarchy of features beneficial during the generation phase. By using only 100 labelled data points, we present state of the art log-likelihood performance on permutation-invariant models for MNIST, and an improvement with respect to comparable models on the OMNIGLOT data set.
While the main focus of this paper is semi-supervised generation, we also show that the same model is able to achieve competitive semi-supervised classification results.

%
\section{Variational Auto-encoders}
\begin{figure}[t]
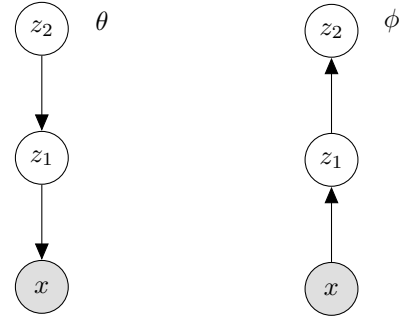

\centering
\def\col{blue}
\begin{subfigure}{0.22\textwidth}
\centering	  
      \tikz{
     \node[obs] (x) {$x$};%
     \node[latent,above=of x] (z1) {$z_1$}; %
     \node[latent,above=of z1] (z2) {$z_2$}; %
     \node[above=of z2, xshift=.8cm, yshift=-1.5cm] (theta) {$\theta$}; %
     \edge[]{z1}{x}
     \edge[]{z2}{z1}
     }
     \caption{Generative model $p_\theta$}\label{fig:vae_gen}
\end{subfigure}
\begin{subfigure}{0.22\textwidth}
\centering	  
      \tikz{
     \node[obs] (x) {$x$};%
     \node[latent,above=of x] (z1) {$z_1$}; %
     \node[latent,above=of z1] (z2) {$z_2$}; %
     \node[above=of z2, xshift=.8cm, yshift=-1.5cm] (phi) {$\phi$}; %
     \edge[]{x}{z1}
     \edge[]{z1}{z2}
     }
     \caption{Inference model $q_\phi$}
     \label{fig:vae_inf}
\end{subfigure}
\caption{Generative model and inference model of a Variational Auto-Encoder with two stochastic layers.}
\label{fig:vae}
\end{figure}

A \textit{Variational Auto-Encoder (VAE)} \citep{Kingma13,Rezende14} defines a deep generative model for data $x$ that depends on latent variable $z$ or a hierarchy of latent variables, e.g.\ $z = [z_1, z_2]$, see Figure \ref{fig:vae_gen} for a graphical representation.
The joint distribution of the two-level generative model is given by
\begin{align*}
p_\theta(x,z_1,z_2)=p_\theta(x|z_1)p_\theta(z_1|z_2)p(z_2) \ ,
\end{align*}
where
\begin{align*}
p_\theta(z_1|z_2) & =\mathcal{N}(z_1; \mu_\theta^1(z_2), \sigma_\theta^1(z_2)) \\
 p(z_2) & =\mathcal{N}(z_2; 0, I)
\end{align*}
are Gaussian distributions with a diagonal covariance matrix and $p_\theta(x|z_1)$ is typically a parameterized Gaussian (continuous data) or Bernoulli distribution (binary data).
The probability distributions of the generative model of a VAE are parameterized using deep neural networks whose parameters are denoted by $\theta$.
Training is performed by optimizing the \textit{Evidence Lower Bound (ELBO)}, a lower bound to the intractable log-likelihood $\log p_\theta(x)$ obtained using Jensen's inequality:
\begin{align}
\log p_\theta(x) & = \log \iint p_\theta(x,z_1,z_2) \drm z_1 \drm z_2 \nonumber \\
& \geq \mathbb{E}_{q_\phi(z_1,z_2|x)}\left[ \log \frac{p_\theta(x,z_1,z_2)}{q_\phi(z_1,z_2|x)}  \right] = \mathcal{F}(\theta, \phi) \ . \label{eq:elbo_vae}
\end{align}
The introduced variational distribution $q_\phi(z_1,z_2|x)$ is an approximation to the model's posterior distribution $p_\theta(z_1,z_2|x)$, defined with a bottom-up dependency structure where each variable of the model depends on the variable below in the hierarchy:
\begin{align*}
q_\phi(z_1,z_2|x)&=q_\phi(z_1|x)q_\phi(z_2|z_1) \\
q_\phi(z_1|x) & = \mathcal{N}(z_1; \mu_\phi^1(x), \sigma_\phi^1(x)) \\
q_\phi(z_2|z_1)& = \mathcal{N}(z_2; \mu_\phi^2(z_1), \sigma_\phi^2(z_1)) \ .
\end{align*}
Similar to the generative model, the mean and diagonal covariance of both Gaussian distributions defining the inference network $q_\phi$ are parameterized with deep neural networks that depend on parameters $\phi$, see Figure \ref{fig:vae_inf} for a graphical representation.

We can learn the parameters $\theta$ and $\phi$ by jointly maximizing the ELBO $\mathcal{F}(\theta, \phi)$ in \eqref{eq:elbo_vae} with stochastic gradient ascent, using Monte Carlo integration to approximate the intractable expectations and computing low variance gradients with the reparameterization trick \citep{Kingma13,Rezende14}.

\paragraph{Inactive stochastic units}

A common problem encountered when training VAEs with bottom-up inference networks is given by the so called \textit{inactive units} in the higher layers of stochastic variables \citep{Burda15,Sonderby2016}. 
In a 2-layer model for example, VAEs often learn $q_\phi(z_2|z_1)=p(z_2)=\mathcal{N}(z_2; 0, I)$, i.e. the variational approximation of $z_2$ uses no information coming from the data point $x$ through $z_1$.
 If we rewrite the ELBO in \eqref{eq:elbo_vae} as
\begin{align*}
\mathcal{F}(\theta, \phi) & = \mathbb{E}_{q_\phi(z_1,z_2|x)}\left[ \log \frac{p_\theta(x,z_1|z_2)}{q_\phi(z_1|x)}  \right] - \\
& \qquad  \mathbb{E}_{q_\phi(z_1|x)}\left[ KL \left[ q_\phi(z_2|z_1) || p(z_2) \right]  \right]
\end{align*}
we can see that $q_\phi(z_2|z_1)=p(z_2)$ represents a local maxima of our optimization problem where the KL-divergence term is set to zero and the information flows by first 
sampling in $\widetilde{z_1} \sim q_\phi(z_1|x)$ and then computing $p_\theta(x|\widetilde{z_1})$ (and is therefore independent from $z_2$). Several techniques have been developed in the literature to mitigate
the problem of inactive units, among which we find annealing of the KL term \citep{Bowman2015,Sonderby2016} or the use of free bits \citep{Kingma2016}.

Using ideas from \citet{Chen2017}, we notice that the inactive units in a VAE with 2 layers of stochastic units can be justified not only as a poor local maxima, but also from the modelling point of view. 
\citet{Chen2017} give a \textit{bits-back coding} interpretation of Variational Inference for a generative model of the form $p(x,z)=p(x|z)p(z)$, with data $x$ and stochastic units $z$. 
The paper shows that if the decoder $p(x|z)$ is powerful enough to explain most of the structure in the data (e.g. an autoregressive decoder), then it will be convenient for the model to set $q(z|x)=p(z)$ not to incur in an extra optimization cost of $KL[q(z|x)||p(z|x)]$. 
The inactive $z_2$ units in a 2-layer VAE can therefore be seen as caused by the flexible distribution $p_\theta(x,z_1|z_2)$ that is able to explain most of the structure in the data without using information from $z_2$. 
By making $q_\phi(z_2|z_1)=p(z_2)$, the model can avoid the extra cost of $KL \left[ q_\phi(z_2|x) || p_\theta(z_2|x) \right]$. A more detailed discussion on the topic can be found in Appendix \ref{app:a}.

It is now clear that if we want a VAE to exploit the power of additional stochastic layers we need to define it so that the benefits of encoding meaningful information in $z_2$ is greater than the cost $KL \left[ q_\phi(z_2|x) || p_\theta(z_2|x) \right]$ that the model has to pay. 
As we will discuss below, we will achieve this by aiding the generative model to do representation learning.

\section{\namep}
Hierarchical models parameterized by deep neural networks have the ability to represent very flexible distributions. 
However, in the previous section we have seen that the units in the higher stochastic layers of a VAE often become inactive.
We will show that we can help the model to exploit the higher stochastic layers by explicitly encoding a useful representation, i.e. the ability to model the natural clustering of the data \citep{Bengio2013a}, which will also be needed for semi-supervised generation.

\begin{figure}[t]
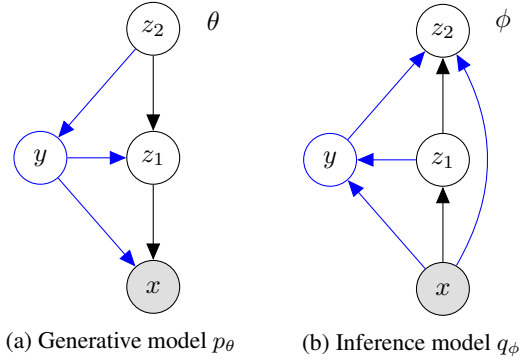

\centering
\def\col{blue}
\begin{subfigure}{0.22\textwidth}
\centering	  
      \tikz{
     \node[obs] (x) {$x$};%
     \node[latent, above=of x, xshift=-1.5cm, draw=\col] (y) {$y$}; %
     \node[latent,above=of x] (z1) {$z_1$}; %
     \node[latent,above=of z1] (z2) {$z_2$}; %
     \node[above=of z2, xshift=.8cm, yshift=-1.5cm] (theta) {$\theta$}; %
     \edge[]{z1}{x}
     \edge[\col]{z2}{y}
     \edge[]{z2}{z1}
     \edge[\col]{y}{x}
     \edge[\col]{y}{z1}
     }
     \caption{Generative model $p_\theta$}
\end{subfigure}
\begin{subfigure}{0.22\textwidth}
\centering	  
      \tikz{
     \node[obs] (x) {$x$};%
     \node[latent, above=of x, xshift=-1.5cm, draw=\col] (y) {$y$}; %
     \node[latent,above=of x] (z1) {$z_1$}; %
     \node[latent,above=of z1] (z2) {$z_2$}; %
     \node[above=of z2, xshift=.8cm, yshift=-1.5cm] (phi) {$\phi$}; %
     \edge[]{x}{z1}
     \edge[]{z1}{z2}
     \edge[\col]{x}{y}
     \edge[\col]{y}{z2}
     \edge[\col]{z1}{y}
     \edge[\col, bend right]{x}{z2}
     }
     \caption{Inference model $q_\phi$}
\end{subfigure}
\caption{Generative model and inference model of a \nm with two stochastic layers (black and blue lines). The black lines only represent a standard VAE.}
\label{fig:simple_graphical_model2}
\end{figure}

We favor the flow of higher-level global information through $z_2$ by extending the generative model of a VAE with a discrete variable $y$ representing the choice of one out of $K$ different clusters in the data.  
The joint distribution $p_\theta(x,z_1,z_2)$ is computed by marginalizing over $y$:
\begin{align*}
p_\theta(x,z_1,z_2) & = \sum_y p_\theta(x,y,z_1,z_2) \\
& = \sum_y p_\theta(x|y,z_1)p_\theta(z_1|y,z_2)p_\theta(y|z_2)p(z_2) \ .
\end{align*}
We call this model \textit{Cluster-aware Generative Model (CaGeM)}, see Figure \ref{fig:simple_graphical_model2} for a graphical representation.
The introduced categorical distribution $p_\theta(y|z_2)=\text{Cat}(y;\pi_\theta(z_2))$ ($\pi_\theta$ represents the class distribution) depends solely on $z_2$, that needs therefore to stay active for the model to be able to represent clusters in the data. 
We further add the dependence of $z_1$ and $x$ on $y$, so that they can now both also represent cluster-dependent information.

\subsection{Inference}
As done for the VAE in \eqref{eq:elbo_vae}, we can derive the ELBO for \nm by maximizing the log-likelihood 
\begin{align*}
\log p_\theta(x) & = \log \iint p_\theta(x,z_1,z_2) \drm z_1 \drm z_2 \\
& = \log \iint \sum_y p_\theta(x,y,z_1,z_2) \drm z_1 \drm z_2 \\
& \geq \mathbb{E}_{q_\phi(y,z_1,z_2|x)}\left[ \log \frac{p_\theta(x,y,z_1,z_2)}{q_\phi(y,z_1,z_2|x)}  \right]  \ .
\end{align*}
We define the variational approximation $q_\phi(y,z_1,z_2|x)$ over the latent variables of the model as
\begin{align*}
q_\phi(y,z_1,z_2|x) = q_\phi(z_2|x,y,z_1)q_\phi(y|z_1,x)q_\phi(z_1|x) \ ,
\end{align*}
where
\begin{align*}
q_\phi(z_1|x) & = \mathcal{N}(z_1; \mu_\phi^1(x), \sigma_\phi^1(x)) \\
q_\phi(z_2|x,y,z_1) & = \mathcal{N}(z_2; \mu_\phi^2(y,z_1), \sigma_\phi^2(y,z_1))\\
q_\phi(y|z_1,x) & = \text{Cat}(y;\pi_\phi(z_1,x))
\end{align*}

In the inference network we then reverse all the dependencies among random variables in the generative model (the arrows in the graphical model in Figure \ref{fig:simple_graphical_model2}). 
This results in a \textit{bottom-up} inference network that performs a feature extraction that is fundamental for learning a good representation of the data. 
Starting from the data $x$ we construct higher levels of abstraction, first through the variables $z_1$ and $y$, and finally through the variable $z_2$, that includes the global information used in the generative model. In order to make the higher representation more expressive we add a skip-connection from $x$ to $z_2$, that is however not fundamental to improve the performances of the model.

With this factorization of the variational distribution $q_\phi(y,z_1,z_2|x)$, the ELBO can be written as 
\begin{align*}
\mathcal{F}(\theta, \phi) & = \mathbb{E}_{q_\phi(z_1|x)}\left[ \sum_y q_\phi(y|z_1,x) \cdot \right. \\ 
& \qquad \left. \cdot \mathbb{E}_{q_\phi(z_2|x,y,z_1)}\left[  \log \frac{p_\theta(x,y,z_1,z_2)}{q_\phi(y,z_1,z_2|x)} \right] \right]  \ .
\end{align*}

We maximize $\mathcal{F}(\theta, \phi)$ by jointly updating, with stochastic gradient ascent, the parameters $\theta$ of the generative model and $\phi$ of the variational approximation. 
When computing the gradients, the summation over $y$ is performed analytically, whereas the intractable expectations over $z_1$ and $z_2$ are approximated by sampling. We use the reparameterization trick to reduce the variance of the stochastic gradients.

\section{Semi-Supervised Generation with CaGeM}
In some applications we may have class label information for some of the data points in the training set. In the following we will show that \nm provides a natural way to exploit additional labelled data to improve the performance of the generative model.
Notice that this \textit{semi-supervised generation} approach differs from the more traditional \textit{semi-supervised classification} task that uses unlabelled data to improve classification accuracies \citep{Kingma14,Maaloe2016,Salimans2016}. 
In our case in fact, it is the labelled data that supports the generative task. Nevertheless, we will see in our experiment that \nm also leads to competitive semi-supervised classification performances.

To exploit the class information, we first set the number of clusters $K$ equal to the number of classes $C$.
We can now define two classifiers in \nm:
\begin{enumerate}
\item In the inference network we can compute the class probabilities given the data, i.e. $q_\phi(y|x)$, by integrating out the stochastic variables $z_1$ from $q_\phi(y,z_1|x)$
\begin{align*}
q_\phi(y|x) & = \int q_\phi(y,z_1|x) \drm z_1 \\
 & = \int q_\phi(y|z_1,x)q_\phi(z_1|x) \drm z_1
\end{align*}

\item Another set of class-probabilities can be computed using the generative model. Given the posterior distribution $p_\theta(z_2|x)$ we have in fact
\begin{align*}
p_\theta(y|x) = \int p_\theta(y|z_2)p_\theta(z_2|x) \drm z_2 \ .
\end{align*}
The posterior over $z_2$ is intractable, but we can approximate it using the variational approximation $q_\phi(z_2|x)$, that is obtained by marginalizing out $y$ and the variable $z_1$ in the joint distribution $q_\phi(y,z_1,z_2|x)$:
\begin{align*}
p_\theta(y|x) & \approx \int p_\theta(y|z_2)q_\phi(z_2|x) \drm z_2 \\
& = \int p_\theta(y|z_2) \left( \int \sum_{\widetilde{y}} q_\phi(z_2|x,\widetilde{y},z_1) \cdot \right. \\ & \qquad \qquad \cdot q_\phi(\widetilde{y}|z_1,x)q_\phi(z_1|x) \drm z_1 \Bigg) \drm z_2 \ . \\ 
\end{align*}
While for the labels $\widetilde{y}$ the summation can be carried out analytically, for the variable $z_1$ and $z_2$ we use Monte Carlo integration.
For each of the $C$ classes we will then obtain a different $z_2^c$ sample $(c=1, \dots C)$ with a corresponding weight given by $q_\phi(\widetilde{y}^c|z_1,x)$. This therefore resembles a \textit{cascade} of classifiers, as the class probabilities of the $p_\theta(y|x)$ classifier will depend on the probabilities of the classifier $q_\phi(y|z_1,x)$ in the inference model.
\end{enumerate}

As our main goal is to learn representations that will lead to good generative performance, we interpret the classification of the additional labelled data as a secondary task that aids in learning a $z_2$ feature space that can be easily separated into clusters. 
We can then see this as a form of \textit{semi-supervised clustering} \citep{Basu2002}, where we know that some data points belong to the same cluster and we are free to learn a data manifold that makes this possible.

The optimal features for the classification task could be very different from the representations learned for the generative task. This is why it is important not to update the parameters of the distributions over $z_1$, $z_2$ and $x$, in both generative model and inference model, using labelled data information. If this is not done carefully, the model could be prone to overfitting towards the labelled data. We define as $\theta_y$ the subset of $\theta$ containing the parameters in $p_\theta(y|z_2)$, and as $\phi_y$ the subset of $\phi$ containing the parameters in $q_\phi(y|z_1,x)$. $\theta_y$ and $\phi_y$ then represent the incoming arrows to $y$ in Figure \ref{fig:simple_graphical_model2}. 
We update the parameters $\theta$ and $\phi$ jointly by maximizing the new objective
\begin{align*}
\mathcal{I} = \sum_{\{x_u\}} \mathcal{{F}}(\theta, \phi) - \alpha \left(\sum_{\{x_l,y_l\}} \left(\mathcal{H}_p(\theta_y, \phi_y) + \mathcal{H}_q(\phi_y) \right) \right)
\end{align*}
where $\{x_u\}$ is the set of unlabelled training points, $\{x_l,y_l\}$ is the set of labelled ones, and $\mathcal{H}_p$ and $\mathcal{H}_q$ are the standard categorical cross-entropies for the $p_\theta(y|x)$ and $q_\phi(y|x)$ classifiers respectively. Notice that we consider the cross-entropies only a function of $\theta_y$ and $\phi_y$, meaning that the gradients of the cross-entropies with respect to the parameters of the distributions over $z_1$, $z_2$ and $x$ will be 0, and will not depend on the labelled data (as needed when learning meaningful representations of the data to be used for the generative task).
To match the relative magnitudes between the ELBO $\mathcal{{F}}(\theta, \phi)$ and the two cross-entropies we set $\alpha=\beta \frac{N_u+N_l}{N_l}$ as done in \citep{Kingma14,Maaloe2016}, where $N_u$ and $N_l$ are the numbers of unlabelled and labelled data points, and $\beta$ is a scaling constant. 

\begin{figure*}[t]
\centering
\includegraphics[width=1.\textwidth]{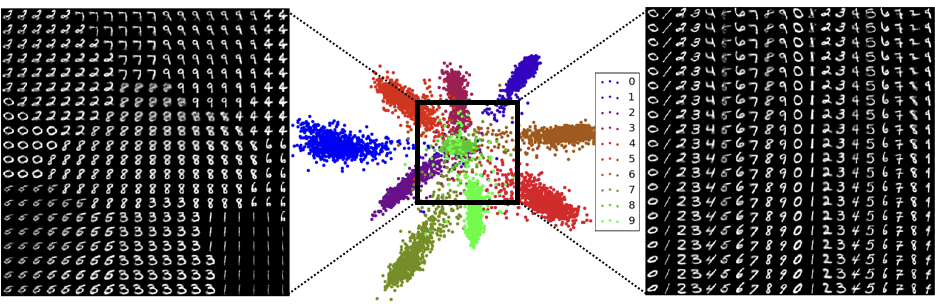}
\caption{Visualizations from CaGeM-100 with a 2-dimensional $z_2$ space. The middle plot shows the latent space, from which we generate random samples (left) and class conditional random samples (right) with a mesh grid (black bounding box). The relative placement of the samples in the scatter plot corresponds to a digit in the mesh grid.}
\label{fig:mesh_plots}
\end{figure*}

\section{Experiments}
We evaluate CaGeM by computing the generative log-likelihood performance on MNIST and OMNIGLOT \citep{Lake2013} datasets. The model is parameterized by feed-forward neural networks ($\mathtt{NN}$) and linear layers ($\mathtt{Linear}$), so that, for Gaussian outputs, each collection of incoming edges to a node in Figure \ref{fig:simple_graphical_model2} is defined as:
\begin{align}
d = \mathtt{NN}(x) \nonumber  \qquad
\mu = \mathtt{Linear}(d)\nonumber   \qquad
\log \sigma = \mathtt{Linear}(d)\ . \ 
\end{align}
For Bernoulli distributed outputs we simply define a feed-forward neural network with a sigmoid activation function for the output. Between dense layers we use the rectified linear unit as non-linearity and batch-normalization \citep{Ioffe2015}. We only collect statistics for the batch-normalization during unlabelled inference. For the log-likelihood experiments we apply temperature on the $KL$-terms during the first 100 epochs of training \citep{Bowman2015,Sonderby2016}. The stochastic layers are defined with $\mathtt{dim}(z_1)=64$, $\mathtt{dim}(z_2)=32$ and 2-layered neural feed-forward networks with respectively 1024 and 512 units in each layer. Training is performed using the Adam optimizer \citep{Kingma14a} with an initial learning rate of $0.001$ and annealing it by $.75$ every $50$ epochs. The experiments are implemented with Theano \citep{Bastien12}, Lasagne \citep{Dieleman15} and Parmesan\footnote{A variational repository named parmesan on Github.}.

For both datasets we report unsupervised and semi-supervised permutation invariant log-likelihood performance and for MNIST we also report semi-supervised classification errors. The input data is dynamically binarized and the ELBO is evaluated by taking 5000 importance-weighted (IW) samples, denoted $\mathcal{F}_{5000}$. We evaluate the performance of CaGeM with different numbers of labelled samples referred to as CaGeM-$\#\mathtt{labels}$. When used, the labelled data is randomly sampled evenly across the class distribution. All experiments across datasets are run with the same architecture.

\section{Results}
Table \ref{table:unsupervised_benchmarks_mnist} shows the generative log-likelihood performances of different variants of CaGeM on the MNIST data set. 
We can see that the more labelled samples we use, the better the generative performance will be.
Even though the results are not directly comparable, since CaGeM exploits a small fraction supervised information, we find that using only 100 labelled samples (10 samples per class), CaGeM-100 model achieves state of the art log-likelihood performance on permutation invariant MNIST with a simple 2-layered model. We also trained a ADGM-100 from \citet{Maaloe2016}\footnote{We used the code supplied in the repository named auxiliary-deep-generative-models on Github.} in order to make a fair comparison on generative log-likelihood in a semi-supervised setting and reached a performance of $-86.06\ \text{nats}$. This indicates that models that are highly optimized for improving semi-supervised classification accuracy may be a suboptimal choice for generative modeling. 

\begin{table}[t]
\begin{center}
\begin{small}
\begin{sc}
\resizebox{\columnwidth}{!}{%
\begin{tabular}{l c}
 & $\leq \log p(x)$ \\
\hline
\abovespace
\underline{Non-Permutation Invariant}  & \\
\abovespace 
DRAW+VGP \citep{Tran2016} & $-79.88$ \\
IAF VAE \citep{Kingma2016} & $-79.10$\\
VLAE  \citep{Chen2017} & $-78.53$\\
\abovespace
\underline{Permutation Invariant}  & \\
\abovespace 
AVAE, L=2, IW=1 \citep{Maaloe2016} & $-82.97$ \\
IWAE, L=2, IW=50 \citep{Burda15} & $-82.90$\\
LVAE, L=5, IW=10 \citep{Sonderby2016} & $-81.74$\\
VAE+VGP, L=2 \citep{Tran2016} & $-81.32$\\
DVAE \citep{Rolfe2017} & $-80.04$\\
 \abovespace 
 CaGeM-0, L=2, IW=1, K=20 & $-82.18$ \\
 CaGeM-0, L=2, IW=1, K=10 & $-81.60$ \\
 CaGeM-20, L=2, IW=1 & $-81.47$ \\
 CaGeM-50, L=2, IW=1 & $-80.49$ \\
 CaGeM-100, L=2, IW=1 & $-79.38$ \\
\hline
\end{tabular}%
}
\end{sc}
\end{small}
\end{center}
\vskip -0.1in
\caption{Test log-likelihood for permutation invariant and non-permutation invariant MNIST. L, IW and K denotes the number of stochastic layers (if it is translatable to the VAE), the number of importance weighted samples used during inference, and the number of predefined clusters used.}
\label{table:unsupervised_benchmarks_mnist}
\vspace{-2mm}
\end{table}

CaGeM could further benefit from the usage of non-permutation invariant architectures suited for image data, such as the autoregressive decoders used by IAF VAE \citep{Kingma2016} and VLAE \citep{Chen2017}. 
The fully unsupervised CaGeM-0 results show that by defining clusters in the higher stochastic units, we achieve better performances than the closely related IWAE \citep{Burda15} and LVAE \citep{Sonderby2016} models.
It is finally interesting to see from Table \ref{table:unsupervised_benchmarks_mnist} that CaGeM-0 performs well even when the number of clusters are different from the number of classes in the labelled data set.

\begin{figure}[t]
\centering
\includegraphics[width=.5\textwidth]{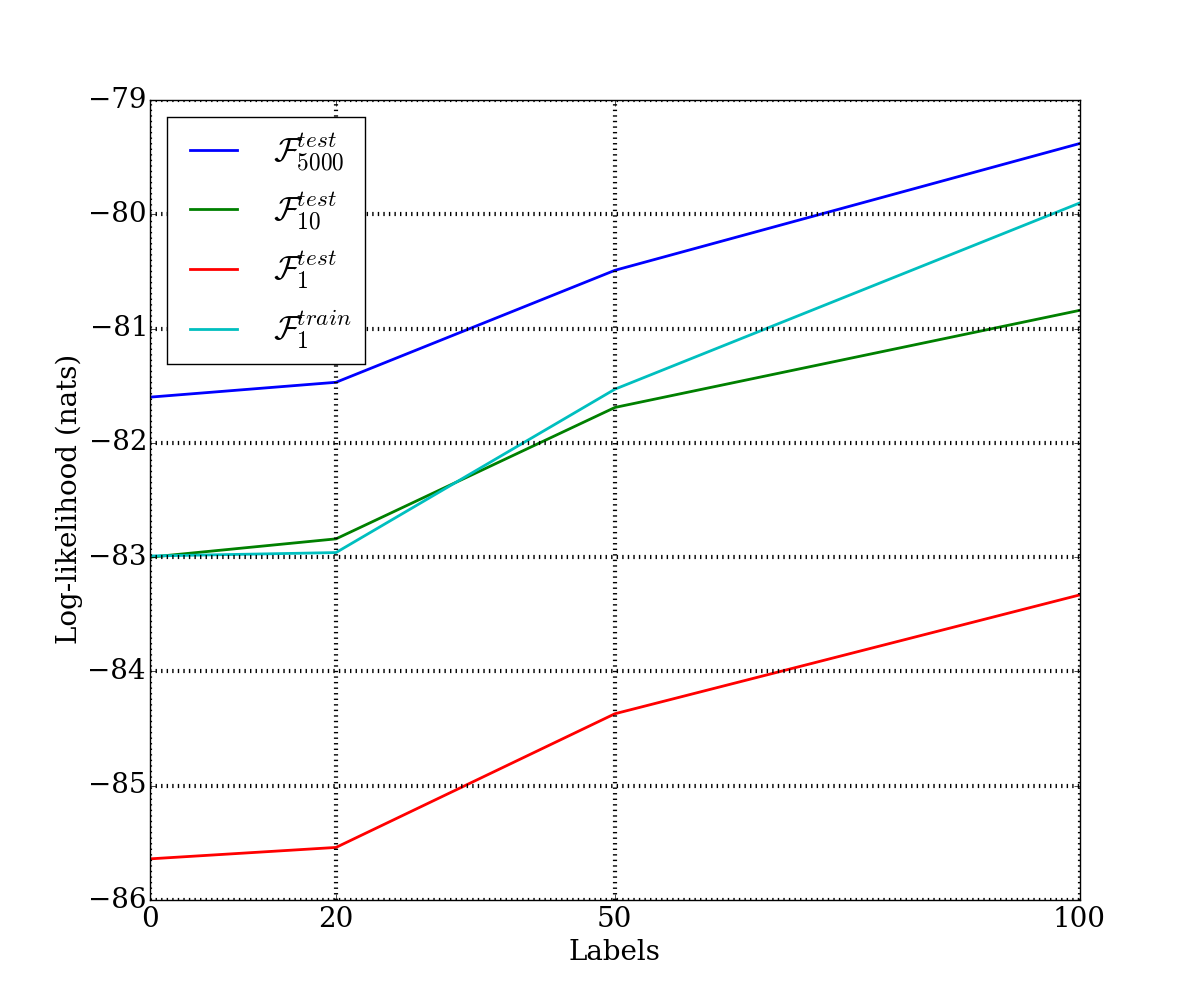}
\caption{Log-likelihood scores for CaGeM on MNIST with 0, 20, 50 and 100 labels with 1, 10 and 5000 IW samples.}
\label{fig:graph_compare}
\end{figure}

In Figure \ref{fig:graph_compare} we show in detail how the performance of CaGeM increases as we add more labelled data points. We can also see that the ELBO $\mathcal{F}^{test}_1$ tightens when adding more labelled information, as compared to $\mathcal{F}^{\text{LVAE}}_1= -85.23$ and $\mathcal{F}^{\text{VAE}}_1= -87.49$ \citep{Sonderby2016}.

\begin{figure}[!ht]
\centering
\includegraphics[width=.4\textwidth]{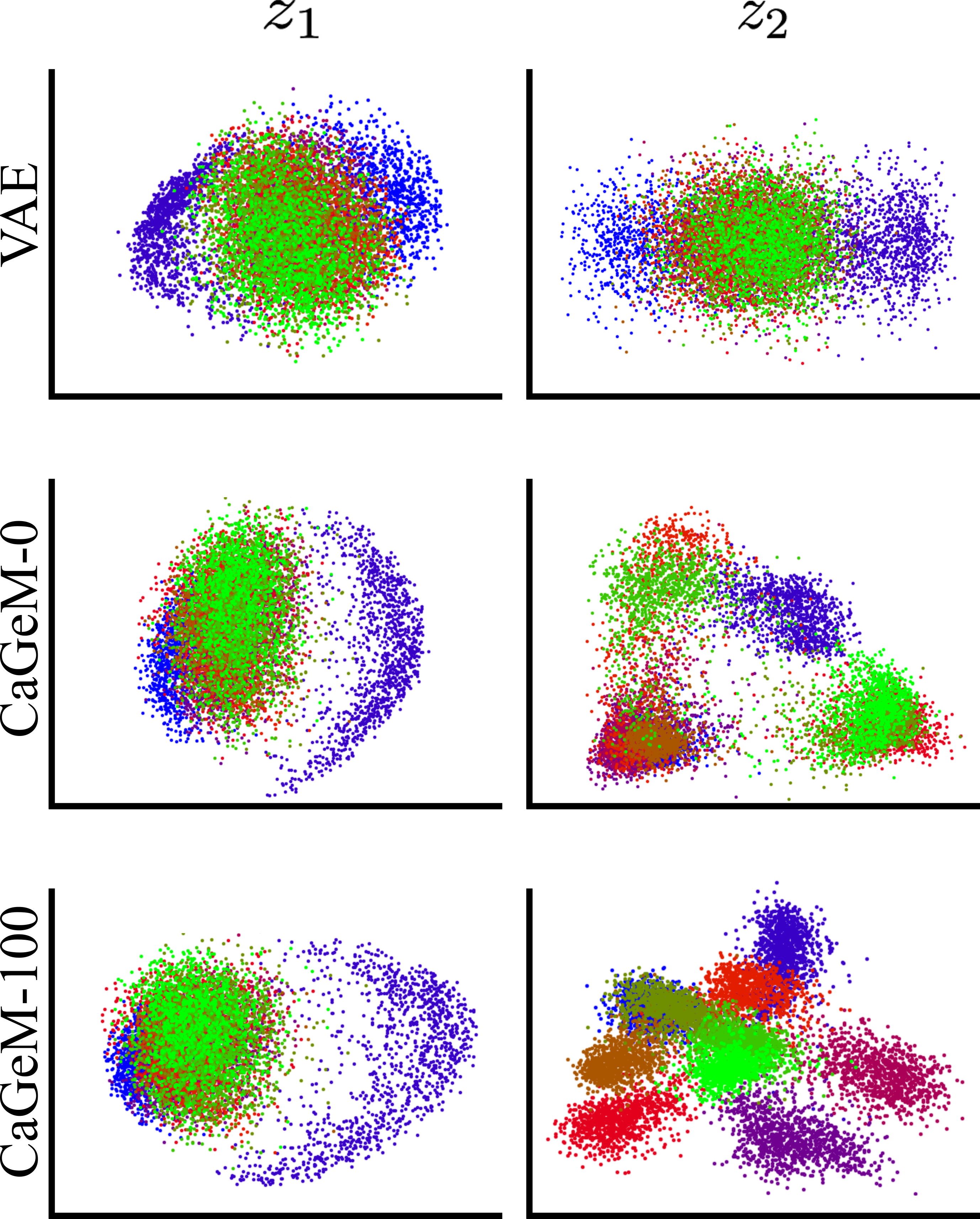}
\caption{PCA plots of the stochastic units $z_1$ and $z_2$ in a 2-layered model trained on MNIST. The colors corresponds to the true labels.}
\label{fig:pca_stochastic_units}
\end{figure}

The PCA plots of the $z_2$ variable of a VAE, CaGeM-0 and CaGeM-100 are shown in Figure \ref{fig:pca_stochastic_units} . We see how CaGeMs encode clustered information into the higher stochastic layer. Since CaGeM-0 is unsupervised, it forms less class-dependent clusters compared to the semi-supervised CaGeM-100, that fits its $z_2$ latent space into 10 nicely separated clusters. Regardless of the labelled information added during inference, CaGeM manages to activate a high amount of units, as for CaGeM we obtain $KL[q_{\phi}(z_2|x,y)||p(z_2)]\approx 17\ \text{nats}$, while a LVAE with 2 stochastic layers obtains $ \approx 9 \ \text{nats}$.

The generative model in CaGeM enables both random samples, by sampling the class variable $y \sim p_\theta(y|z_2)$ and feeding it to $p_\theta(x|z_1, y)$, and class conditional samples by fixing $y$. Figure \ref{fig:mesh_plots} shows the generation of MNIST digits from CaGeM-100 with $\mathtt{dim}(z_2)=2$. The images are generated by applying a linearly spaced mesh grid within the latent space $z_2$ and performing random generations (left) and conditional generations (right). When generating samples in CaGeM, it is clear how the latent units $z_1$ and $z_2$ capture different modalities within the true data distribution, namely style and class. 

Regardless of the fact that CaGeM was designed to optimize the semi-supervised generation task, the model can also be used for classification by using the classifier $p_\theta(y|x)$. In Table \ref{table:semisupervised} we show that the semi-supervised classification accuracies obtained with CaGeM are comparable to the performance of GANs \cite{Salimans2016}.

\begin{table*}[!ht]
\begin{center}
\begin{small}
\begin{sc}
\begin{tabular}{llll}
\abovespace
  Labels & $20$ & $50$ & $100$ \\
\hline
\abovespace 
M1+M2 \citep{Kingma14} & - & - & $3.33$\% ($\pm 0.14$) \\
VAT \citep{Miyato15} & - & - & $2.12$\% \\
CatGAN \citep{Springenberg2015} & - & - & $1.91$\% ($\pm 0.1$) \\
SDGM \citep{Maaloe2016} & - & - & $1.32$\% ($\pm 0.07$) \\
Ladder Network \citep{Rasmus15} & - & - & $1.06$\% ($\pm 0.37$)\\ 
ADGM \citep{Maaloe2016} & - & - & $0.96$\% ($\pm 0.02$) \\
Imp. GAN \citep{Salimans2016} & $16.77$\% ($\pm 4.52$) & $2.21$\% ($\pm 1.36$) & $0.93$\% ($\pm 0.65$) \\
\abovespace
CaGeM & $15.86$\% & $2.42$\% & $1.16$\%\\
\hline
\end{tabular}
\end{sc}
\end{small}
\end{center}
\vskip -0.1in
\caption{Semi-supervised test error \% benchmarks on MNIST for 20, 50, and 100 randomly chosen and evenly distributed labelled samples. Each experiment was run 3 times with different labelled subsets and the reported accuracy is the mean value.}\label{table:semisupervised}
\vspace{-2mm}
\end{table*}

\begin{figure}[!h]
\centering
\includegraphics[width=.47\textwidth]{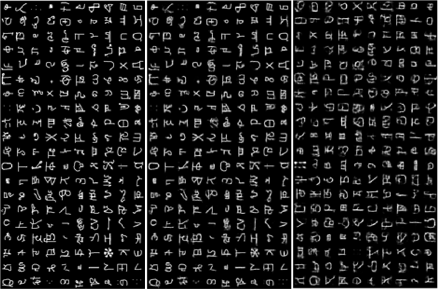}
\caption{Generations from CaGeM-500. (left) The input images, (middle) the reconstructions, and (right) random samples from $z_2$.}
\label{fig:omniglot}
\end{figure}

The OMNIGLOT dataset consists of 50 different alphabets of handwritten characters, where each character is sparsely represented. In this task we use the alphabets as the cluster information, so that the $z_2$ representation should divide correspondingly. From Table \ref{table:unsupervised_benchmarks_omniglot} we see an improvement over other comparable VAE architectures (VAE, IWAE and LVAE), however, the performance is far from the once reported from the auto-regressive models \citep{Kingma2016,Chen2017}. This indicates that the alphabet information is not as strong as for a dataset like MNIST. This is also indicated from the accuracy of CaGeM-500, reaching a performance of $\approx 24 \%$. Samples from the model can be found in Figure \ref{fig:omniglot}.
	
\begin{table}[!h]
\vspace*{-0.6cm}
\begin{center}
\begin{small}
\begin{sc}
\resizebox{\columnwidth}{!}{%
\begin{tabular}{l c}
 & $\leq \log p(x)$ \\
\hline
\abovespace
VAE,  L=2, IW=50 \citep{Burda15} & $-106.30$\\
IWAE, L=2, IW=50 \citep{Burda15} & $-103.38$\\
LVAE, L=5, FT, IW=10 \citep{Sonderby2016} & $-102.11$\\
RBM \citep{Burda15a} & $-100.46$\\
DBN \citep{Burda15a} & $-100.45$\\
DVAE \citep{Rolfe2017} & $-97.43$\\
 \abovespace 
CaGeM-500, L=2, IW=1 & $-100.86$ \\
\hline
\end{tabular}%
}
\end{sc}
\end{small}
\end{center}
\vskip -0.1in
\caption{Generative test log-likelihood for permutation invariant OMNIGLOT.}
\label{table:unsupervised_benchmarks_omniglot}
\vspace{-2mm}
\end{table}

\vspace*{-.3cm}
\section{Discussion}
As we have seen from our experiments, CaGeM offers a way to exploit the added flexibility of a second layer of stochastic units, 
that stays active as the modeling performances can greatly benefit from capturing the natural clustering of the data.
Other recent works have presented alternative methods to mitigate the problem of inactive units when training flexible models defined by a hierarchy of stochastic layers. 
\citet{Burda15} used importance samples to improve the tightness of the ELBO, and showed that this new training objective helped in activating the units of a 2-layer VAE.
\citet{Sonderby2016} trained Ladder Variational Autoencoders (LVAE) composed of up to 5 layers of stochastic units, using a top-down inference network that forces the information to flow in the higher stochastic layers. 
Contrarily to the bottom-up inference network of CaGeM, the top-down approach used in LVAEs does not enforce a clear separation between the role of each stochastic unit, as proven by the fact that all of them encode some class information.
Longer hierarchies of stochastic units unrolled in time can be found in the sequential setting \citep{Krishnan2015,Fraccaro2016}. 
In these applications the problem of inactive stochastic units appears when using powerful autoregressive decoders \citep{Fraccaro2016,Chen2017}, but is mitigated by the fact that new data information enters the model at each time step. 

The discrete variable $y$ of CaGeM was introduced to be able to define a better learnable representation of the data, that helps in activating the higher stochastic layer.
The combination of discrete and continuous variables for deep generative models was also recently explored by several authors.
\citet{Maddison2016,Jang2016} used a continuous relaxation of the discrete variables, that makes it possible to efficiently train the model using stochastic backpropagation. 
The introduced Gumbel-Softmax variables allow to sacrifice log-likelihood performances to avoid the computationally expensive integration over $y$.
\citet{Rolfe2017} presents a new class of probabilistic models that combines an undirected component consisting of a bipartite Boltzmann machine with binary units and a directed component with multiple layers of continuous variables.

Traditionally, semi-supervised learning applications of deep generative models such as Variational Auto-encoders and Generative Adversarial Networks \citep{Goodfellow2014} have shown that, 
whenever only a small fraction of labelled data is available, the supervised classification task can benefit from additional unlabelled data \citep{Kingma14,Maaloe2016,Salimans2016}.
In this work we consider the semi-supervised problem from a different perspective, and show that the generative task of CaGeM can benefit from additional labelled data. 
As a by-product of our model however, we also obtain competitive semi-supervised classification results, meaning that CaGeM is able to share statistical strength between the generative and classification tasks.

When modeling natural images, the performance of CaGeM could be further improved using more powerful autoregressive decoders such as the ones in \citep{Gulrajani2016,Chen2017}. 
Also, an even more flexible variational approximation could be obtained using auxiliary variables \cite{Ranganath2015,Maaloe2016} or normalizing flows \cite{Rezende2015,Kingma2016}.

\section{Conclusion}
In this work we have shown how to perform semi-supervised generation with CaGeM.
We showed that CaGeM improves the generative log-likelihood performance over similar deep generative approaches by creating clusters for the data in its higher latent representations using unlabelled information. CaGeM also provides a natural way to refine the clusters using additional labelled information to further improve its modelling power.

\appendix 
\section{The Problem of Inactive Units}
\label{app:a}
First consider a model $p(x)$ without latent units. We consider the asymptotic average properties, so we take the expectation of the log-likelihood over the (unknown) data distribution $p_{\rm data}(x)$:
\begin{align}
\mathbb{E}_{p_{\rm data}(x)} \left[ \log p(x) \right] & = 
\mathbb{E}_{p_{\rm data}(x)}  \left[ \log \Big(p_{\rm data}(x)\frac{p(x)}{p_{\rm data}(x)}\Big) \right] \nonumber \\
& = 
-\mathcal{H}(p_{\rm data}) - KL(p_{\rm data}(x)||p(x)) \ , \nonumber 
\end{align}
where $\mathcal{H}(p) = - \mathbb{E}_{p(x)} \left[ \log  p(x) \right]$ is the entropy of the distribution and $KL(\cdot||\cdot)$ is the KL-divergence. The expected log-likelihood is then simply the baseline entropy of the data generating distribution minus the deviation between the data generating distribution and our model for the distribution.

For the latent variable model $p_{\rm lat}(x) = \int p(x|z) p(z) dz$ the log-likelihood bound is: 
$$
\log p_{\rm lat}(x) \geq \mathbb{E}_{q(z|x)} \left[ \log \frac{p(x|z) p(z)}{q(z|x)}\right] \ . 
$$
We take the expectation over the data generating distribution and apply the same steps as above
\begin{align}
\mathbb{E}_{p_{\rm data}(x)} & \left[ \log p_{\rm lat}(x) \right]  \geq \mathbb{E}_{p_{\rm data}(x)} \mathbb{E}_{q(z|x)} \left[ \log \frac{p(x|z) p(z)}{q(z|x)}\right] \nonumber \\
& \quad \quad = -\mathcal{H}(p_{\rm data}) -  KL(p_{\rm data}(x)||p_{\rm lat}(x)) \nonumber \\ 
& \quad \quad \quad \quad \quad - \mathbb{E}_{p_{\rm data}(x)} \left[  KL(q(z|x)||p(z|x)) \right] \ , \nonumber
\end{align}
where $p(z|x) = p(x|z)p(z)/p_{\rm lat}(x)$ is the (intractable) posterior of the latent variable model.
This results shows that we pay an additional price (the last term) for using an approximation to the posterior.

The latent variable model can choose to ignore the latent variables, $p(x|z) = \hat{p}(x)$. When this happens the expression falls back to the log-likelihood without latent variables.  We can therefore get an (intractable) condition for when it is advantageous for the model to use the latent variables:
\begin{align}
\mathbb{E}_{p_{\rm data}(x)} \left[ \log p_{\rm lat}(x)) \right]& > \mathbb{E}_{p_{\rm data}(x)} \left[ \log \hat{p}(x)) \right] + \nonumber \\ 
& \quad \mathbb{E}_{p_{\rm data}(x)} \left[  KL(q(z|x)||p(z|x)) \right]  \ .  \nonumber
\end{align}
The model will use latent variables when the log-likelihood gain is so high that it can compensate for the loss $KL(q(z|x)||p_{\rm lat}(z|x))$ we pay by using an approximate posterior distribution.

The above argument can also be used to understand why it is harder to get additional layers of latent variables to become active. For a two-layer latent variable model $p(x,z_1,z_2) = p(x|z_1)p(z_1|z_2)p(z_2)$ we use a variational distribution $q(z_1,z_2|x)=q(z_2|z_1,x)q(z_1|x)$ and decompose the log likelihood bound using $p(x,z_1,z_2)=p(z_2|z_1,x)p(z_1|x)p_{\rm lat,2}(x)$:
\begin{align}
&\mathbb{E}_{p_{\rm data}(x)} \left[ \log p_{\rm lat,2}(x) \right] \nonumber \\ 
& \quad \geq \mathbb{E}_{p_{\rm data}(x)} \mathbb{E}_{q(z_1,z_2|x)} \left[ \log \frac{p(x|z_1)p(z_1|z_2)p(z_2)}{q(z_1,z_2|x)}\right] \nonumber \\ 
& \quad = -H(p_{\rm data}) -  KL(p_{\rm data}(x)||p_{\rm lat,2}(x)) \nonumber \\
& \quad\quad - \mathbb{E}_{p_{\rm data}(x)} \mathbb{E}_{q(z_1|x)}\left[  KL(q(z_2|z_1,x)||p(z_2|z_1,x)) \right] \nonumber \\
& \quad\quad - \mathbb{E}_{p_{\rm data}(x)} KL(q(z_1|x)||p(z_1|x)) \nonumber \ .
\end{align}

Again this expression falls back to the one-layer model when $p(z_1|z_2)=\hat{p}(z_1)$.
So whether to use the second layer of stochastic units will depend upon the potential diminishing return in terms of log likelihood relative to the extra $KL$-cost from the approximate posterior. 

\section*{Acknowledgements} 
We thank Ulrich Paquet for fruitful feedback. The research was supported by  Danish Innovation Foundation, the NVIDIA Corporation with the donation of TITAN X GPUs. Marco Fraccaro is supported by Microsoft Research through its PhD Scholarship Programme.
\bibliography{icml2017_references}

\begin{thebibliography}{32}
\providecommand{\natexlab}[1]{#1}
\providecommand{\url}[1]{\texttt{#1}}
\expandafter\ifx\csname urlstyle\endcsname\relax
  \providecommand{\doi}[1]{doi: #1}\else
  \providecommand{\doi}{doi: \begingroup \urlstyle{rm}\Url}\fi

\bibitem[Bastien et~al.(2012)Bastien, Lamblin, Pascanu, Bergstra, Goodfellow,
  Bergeron, Bouchard, and Bengio]{Bastien12}
Bastien, Fr{\'{e}}d{\'{e}}ric, Lamblin, Pascal, Pascanu, Razvan, Bergstra,
  James, Goodfellow, Ian~J., Bergeron, Arnaud, Bouchard, Nicolas, and Bengio,
  Yoshua.
\newblock {Theano: new features and speed improvements}.
\newblock In \emph{Deep Learning and Unsupervised Feature Learning, workshop at
  Neural Information Processing Systems}, 2012.

\bibitem[Basu et~al.(2002)Basu, Banerjee, and Mooney]{Basu2002}
Basu, Sugato, Banerjee, Arindam, and Mooney, Raymond~J.
\newblock Semi-supervised clustering by seeding.
\newblock In \emph{Proceedings of the International Conference on Machine
  Learning}, 2002.

\bibitem[Bengio et~al.(2013)Bengio, Courville, and Vincent]{Bengio2013a}
Bengio, Yoshua, Courville, Aaron, and Vincent, Pascal.
\newblock Representation learning: A review and new perspectives.
\newblock \emph{IEEE Transactions on Pattern Analysis and Machine
  Intelligence}, 35\penalty0 (8), 2013.

\bibitem[Bowman et~al.(2015)Bowman, Vilnis, Vinyals, Dai, Jozefowicz, and
  Bengio]{Bowman2015}
Bowman, S.R., Vilnis, L., Vinyals, O., Dai, A.M., Jozefowicz, R., and Bengio,
  S.
\newblock Generating sentences from a continuous space.
\newblock \emph{arXiv preprint arXiv:1511.06349}, 2015.

\bibitem[Burda et~al.(2015{\natexlab{a}})Burda, Grosse, and
  Salakhutdinov]{Burda15}
Burda, Yuri, Grosse, Roger, and Salakhutdinov, Ruslan.
\newblock {Importance Weighted Autoencoders}.
\newblock \emph{arXiv preprint arXiv:1509.00519}, 2015{\natexlab{a}}.

\bibitem[Burda et~al.(2015{\natexlab{b}})Burda, Grosse, and
  Salakhutdinov]{Burda15a}
Burda, Yuri, Grosse, Roger, and Salakhutdinov, Ruslan.
\newblock {Accurate and conservative estimates of mrf log-likelihood using
  reverse annealing}.
\newblock In \emph{Proceedings of the International Conference on Artificial
  Intelligence and Statistics}, 2015{\natexlab{b}}.

\bibitem[Chen et~al.(2017)Chen, Kingma, Salimans, Duan, Dhariwal, Schulman,
  Sutskever, and Abbeel]{Chen2017}
Chen, Xi, Kingma, Diederik~P., Salimans, Tim, Duan, Yan, Dhariwal, Prafulla,
  Schulman, John, Sutskever, Ilya, and Abbeel, Pieter.
\newblock {Variational Lossy Autoencoder}.
\newblock In \emph{International Conference on Learning Representations}, 2017.

\bibitem[Dieleman et~al.(2015)Dieleman, Schlüter, Raffel, Olson, S{\o}nderby,
  Nouri, van~den Oord, and and]{Dieleman15}
Dieleman, Sander, Schlüter, Jan, Raffel, Colin, Olson, Eben, S{\o}nderby,
  S{\o}ren~K, Nouri, Daniel, van~den Oord, Aaron, and and, Eric~Battenberg.
\newblock Lasagne: First release., August 2015.

\bibitem[Fraccaro et~al.(2016)Fraccaro, S{\o}nderby, Paquet, and
  Winther]{Fraccaro2016}
Fraccaro, Marco, S{\o}nderby, S{\o}ren~Kaae, Paquet, Ulrich, and Winther, Ole.
\newblock Sequential neural models with stochastic layers.
\newblock In \emph{Advances in Neural Information Processing Systems}. 2016.

\bibitem[Goodfellow et~al.(2014)Goodfellow, Pouget-Abadie, Mirza, Xu,
  Warde-Farley, Ozair, Courville, and Bengio]{Goodfellow2014}
Goodfellow, Ian, Pouget-Abadie, Jean, Mirza, Mehdi, Xu, Bing, Warde-Farley,
  David, Ozair, Sherjil, Courville, Aaron, and Bengio, Yoshua.
\newblock Generative adversarial nets.
\newblock In \emph{Advances in Neural Information Processing Systems}. 2014.

\bibitem[Gulrajani et~al.(2016)Gulrajani, Kumar, Ahmed, Ali~Taiga, Visin,
  Vazquez, and Courville]{Gulrajani2016}
Gulrajani, Ishaan, Kumar, Kundan, Ahmed, Faruk, Ali~Taiga, Adrien, Visin,
  Francesco, Vazquez, David, and Courville, Aaron.
\newblock {PixelVAE}: A latent variable model for natural images.
\newblock \emph{arXiv e-prints}, 1611.05013, November 2016.

\bibitem[Ioffe \& Szegedy(2015)Ioffe and Szegedy]{Ioffe2015}
Ioffe, Sergey and Szegedy, Christian.
\newblock Batch normalization: Accelerating deep network training by reducing
  internal covariate shift.
\newblock In \emph{Proceedings of International Conference on Machine
  Learning}, 2015.

\bibitem[Jang et~al.(2016)Jang, Gu, and Poole]{Jang2016}
Jang, Eric, Gu, Shixiang, and Poole, Ben.
\newblock Categorical reparameterization with gumbel-softmax.
\newblock \emph{arXiv preprint arXiv:1611.01144}, 2016.

\bibitem[Kingma \& Ba(2014)Kingma and Ba]{Kingma14a}
Kingma, Diederik and Ba, Jimmy.
\newblock {Adam: A Method for Stochastic Optimization}.
\newblock \emph{arXiv preprint arXiv:1412.6980}, 12 2014.

\bibitem[Kingma et~al.(2014)Kingma, Rezende, Mohamed, and Welling]{Kingma14}
Kingma, Diederik~P., Rezende, Danilo~Jimenez, Mohamed, Shakir, and Welling,
  Max.
\newblock {Semi-Supervised Learning with Deep Generative Models}.
\newblock In \emph{Proceedings of the International Conference on Machine
  Learning}, 2014.

\bibitem[Kingma et~al.(2016)Kingma, Salimans, Jozefowicz, Chen, Sutskever, and
  Welling]{Kingma2016}
Kingma, Diederik~P, Salimans, Tim, Jozefowicz, Rafal, Chen, Xi, Sutskever,
  Ilya, and Welling, Max.
\newblock Improved variational inference with inverse autoregressive flow.
\newblock In \emph{Advances in Neural Information Processing Systems}. 2016.

\bibitem[Kingma(2013)]{Kingma13}
Kingma, Diederik P;~Welling, Max.
\newblock {Auto-Encoding Variational Bayes}.
\newblock \emph{arXiv preprint arXiv:1312.6114}, 12 2013.

\bibitem[Krishnan et~al.(2015)Krishnan, Shalit, and Sontag]{Krishnan2015}
Krishnan, Rahul~G, Shalit, Uri, and Sontag, David.
\newblock Deep {K}alman filters.
\newblock \emph{arXiv:1511.05121}, 2015.

\bibitem[Lake et~al.(2013)Lake, Salakhutdinov, and Tenenbaum]{Lake2013}
Lake, Brenden~M, Salakhutdinov, Ruslan~R, and Tenenbaum, Josh.
\newblock One-shot learning by inverting a compositional causal process.
\newblock In \emph{Advances in Neural Information Processing Systems}. 2013.

\bibitem[Maal{\o}e et~al.(2016)Maal{\o}e, S{\o}nderby, S{\o}nderby, and
  Winther]{Maaloe2016}
Maal{\o}e, Lars, S{\o}nderby, Casper~K., S{\o}nderby, S{\o}ren~K., and Winther,
  Ole.
\newblock {Auxiliary Deep Generative Models}.
\newblock In \emph{Proceedings of the International Conference on Machine
  Learning}, 2016.

\bibitem[Maddison et~al.(2016)Maddison, Mnih, and Teh]{Maddison2016}
Maddison, Chris~J., Mnih, Andriy, and Teh, Yee~Whye.
\newblock The concrete distribution: {A} continuous relaxation of discrete
  random variables.
\newblock \emph{arXiv preprint arXiv:1611.00712}, abs/1611.00712, 2016.

\bibitem[Miyato et~al.(2015)Miyato, Maeda, Koyama, Nakae, and Ishii]{Miyato15}
Miyato, Takeru, Maeda, Shin-ichi, Koyama, Masanori, Nakae, Ken, and Ishii,
  Shin.
\newblock {Distributional Smoothing with Virtual Adversarial Training}.
\newblock \emph{arXiv preprint arXiv:1507.00677}, 7 2015.

\bibitem[Ranganath et~al.(2015)Ranganath, Tran, and Blei]{Ranganath2015}
Ranganath, Rajesh, Tran, Dustin, and Blei, David~M.
\newblock Hierarchical variational models.
\newblock \emph{arXiv preprint arXiv:1511.02386}, 11 2015.

\bibitem[Rasmus et~al.(2015)Rasmus, Berglund, Honkala, Valpola, and
  Raiko]{Rasmus15}
Rasmus, Antti, Berglund, Mathias, Honkala, Mikko, Valpola, Harri, and Raiko,
  Tapani.
\newblock Semi-supervised learning with ladder networks.
\newblock In \emph{Advances in Neural Information Processing Systems}, 2015.

\bibitem[Rezende et~al.(2014)Rezende, Mohamed, and Wierstra]{Rezende14}
Rezende, Danilo~J., Mohamed, Shakir, and Wierstra, Daan.
\newblock {Stochastic Backpropagation and Approximate Inference in Deep
  Generative Models}.
\newblock \emph{arXiv preprint arXiv:1401.4082}, 04 2014.

\bibitem[Rezende \& Mohamed(2015)Rezende and Mohamed]{Rezende2015}
Rezende, Danilo~Jimenez and Mohamed, Shakir.
\newblock {Variational Inference with Normalizing Flows}.
\newblock In \emph{Proceedings of the International Conference on Machine
  Learning}, 2015.

\bibitem[Rolfe(2017)]{Rolfe2017}
Rolfe, Jason~Tyler.
\newblock Discrete variational autoencoders.
\newblock In \emph{Proceedings of the International Conference on Learning
  Representations}, 2017.

\bibitem[{Salimans} et~al.(2016){Salimans}, {Goodfellow}, {Zaremba}, {Cheung},
  {Radford}, and {Chen}]{Salimans2016}
{Salimans}, T., {Goodfellow}, I., {Zaremba}, W., {Cheung}, V., {Radford}, A.,
  and {Chen}, X.
\newblock Improved techniques for training gans.
\newblock \emph{arXiv preprint arXiv:1606.03498}, 2016.

\bibitem[S{\o}nderby et~al.(2016)S{\o}nderby, Raiko, Maal{\o}e, S{\o}nderby,
  and Winther]{Sonderby2016}
S{\o}nderby, Casper~Kaae, Raiko, Tapani, Maal{\o}e, Lars, S{\o}nderby,
  S{\o}ren~Kaae, and Winther, Ole.
\newblock Ladder variational autoencoders.
\newblock In \emph{Advances in Neural Information Processing Systems 29}. 2016.

\bibitem[{Springenberg}(2015)]{Springenberg2015}
{Springenberg}, J.T.
\newblock Unsupervised and semi-supervised learning with categorical generative
  adversarial networks.
\newblock \emph{arXiv preprint arXiv:1511.06390}, 2015.

\bibitem[Tenenbaum et~al.(2006)Tenenbaum, Griffiths, and Kemp]{Tenenbaum2006}
Tenenbaum, Joshua~B., Griffiths, Thomas~L., and Kemp, Charles.
\newblock {Theory-based Bayesian models of inductive learning and reasoning.}
\newblock \emph{Trends in cognitive sciences}, 10\penalty0 (7):\penalty0
  309--318, July 2006.

\bibitem[Tran et~al.(2016)Tran, Ranganath, and Blei]{Tran2016}
Tran, Dustin, Ranganath, Rajesh, and Blei, David~M.
\newblock Variational {G}aussian process.
\newblock In \emph{Proceedings of the International Conference on Learning
  Representations}, 2016.

\end{thebibliography}
\bibliographystyle{icml2017}

\end{document}